\documentclass{article}

\usepackage[final]{neurips_2025}

\usepackage[utf8]{inputenc} 
\usepackage[T1]{fontenc}    
\usepackage{hyperref}       
\usepackage{url}            
\usepackage{booktabs}       
\usepackage{amsfonts}       
\usepackage{nicefrac}       
\usepackage{microtype}      
\usepackage{xcolor}         
\usepackage{float}
\usepackage{listings}

\usepackage{graphicx}
\newcommand{\sys}{{\em NetGent}}
\usepackage{xcolor}
\usepackage{threeparttable} 

\usepackage{listings}
\usepackage{xcolor}

\usepackage{tikz}
\usetikzlibrary{arrows.meta,positioning,fit,calc}

\usepackage{xspace}

\newcommand{\smartparagraph}[1]{\noindent{\bf #1}\ }

\usepackage{pifont}

\title{\sys{}: Agent-Based Automation of Network Application Workflows}

\author{%
Jaber Daneshamooz$^{*}$~Eugene Vuong$^{\ddagger}$~Laasya Koduru$^{*}$~Sanjay Chandrasekaran$^{*}$~Arpit Gupta$^{*}$ \\
$^{*}$University of California Santa Barbara \quad $^{\ddagger}$California State University, East Bay \\
}

\makeatletter
\def\@trackname{} 
\makeatother

\begin{document}

\maketitle

\begin{abstract}
We present \sys{}, an AI-agent framework for automating complex application workflows to generate realistic network traffic datasets. Developing generalizable ML models for networking requires data collection from network environments with traffic that results from a diverse set of real-world web applications. However, using existing browser automation tools that are diverse, repeatable, realistic, and efficient remains fragile and costly. 
\sys{} addresses this challenge by allowing users to specify workflows as natural-language rules that define state-dependent actions. These abstract specifications are compiled into nondeterministic finite automata (NFAs), which a state synthesis component translates into reusable, executable code. This design enables deterministic replay, reduces redundant LLM calls through state caching, and adapts quickly when application interfaces change. In experiments, \sys{} automated more than 50+ workflows spanning video-on-demand streaming, live video streaming, video conferencing, social media, and web scraping, producing realistic traffic traces while remaining robust to UI variability. By combining the flexibility of language-based agents with the reliability of compiled execution, \sys{} provides a scalable foundation for generating the diverse, repeatable datasets needed to advance ML in networking.
\end{abstract}

\section{Introduction}

Machine learning for networking has become an increasingly active area of research for tasks such as QoE inference and optimization of networked applications. 
A persistent barrier is access to realistic, labeled application data at scale~\cite{netunicorn}. 
Unlike vision or NLP, networking datasets often \emph{cannot} be scraped or passively collected: they must be \emph{generated} by executing live application workflows (e.g., streaming a video, joining a meeting, browsing social media) so that traffic, logs, and user interactions reflect real deployments. 
Today, researchers commonly rely on browser-automation scripts (e.g., Selenium~\cite{seleniumbase}, PyAutoGUI~\cite{pyautogui}) to repeat experiments and scale data collection. 
However, authoring and maintaining such scripts is long, manual, and brittle, especially for complex, multi-step tasks across diverse sites.

As a result, prior work frequently narrows the scope to a small set of applications or use cases when generating datasets~\cite{netmirco}. Yet building generalizable ML models requires collecting across \emph{many} applications, inputs (e.g., different videos), and network environments. 
To keep up with this demand, data collection pipelines must repeatedly execute the same workflows under varied conditions while remaining robust to evolving user interfaces and behavior~\cite{netreplica}. 

Consider a concrete example: automate Disney+ to open ESPN, select the first video, and move the playback slider to the five-minute mark. 
Even this simple task varies: a user may or may not be logged in; profile selection may be required; the ESPN entry point and page layout change over time; ads or PIN prompts may appear. 
Such variability makes it difficult to design automation that is both robust and repeatable, especially when scaled to thousands or millions of runs across diverse network conditions for ML training and evaluation.

\smartparagraph{Requirements.}
This example illustrates six interrelated requirements for networking data generation: 
(1) \emph{diversity} across applications and platforms; 
(2) \emph{repeatability} so identical inputs yield identical outcomes across many runs and network conditions; 
(3) \emph{complexity} to capture dynamic, non-linear, multi-step interactions; 
(4) \emph{robustness} to survive frequent UI changes; 
(5) \emph{realism} to mimic human behavior and avoid undesired bot detection; and 
(6) \emph{efficiency} to minimize token usage and workflow-generation time. 
Meeting all six simultaneously is non-trivial; improving one dimension often degrades another.

\smartparagraph{Why existing approaches fall short.}
Web/GUI agents and script-based automation each solve a subset of these needs. 
Agentic approaches (e.g., ReAct~\cite{yao2023reactsynergizingreasoningacting}, Reflexion~\cite{shinn2023reflexionlanguageagentsverbal}) emphasize online planning and self-reflection, but incur high token costs per execution and remain unreliable on long horizons—GPT-4 agents achieve $\approx14\%$ success on WebArena versus $\approx78\%$ for humans~\cite{zhou2024webarenarealisticwebenvironment}; Mind2Web~\cite{deng2023mind2webgeneralistagentweb}, VisualWebArena~\cite{koh2024visualwebarenaevaluatingmultimodalagents}, and BrowserGym~\cite{dechezelles2025browsergymecosystemwebagent} further document these gaps. 
Scripted frameworks (Selenium, Playwright, PyAutoGUI) provide efficient replay but are notoriously flaky under UI drift; empirical studies attribute failures to asynchronous waits, DOM instability, and timing issues~\cite{luo2014empirical,thorve2018empirical,bqt}. 
More specialized web scraping tools, such as for the broadband-plan querying tool (BQT~\cite{bqt,bqtplus}), trade generality for robustness, but are still fragile to UI changes. 
None of these simultaneously delivers diversity, repeatability at scale, robustness, realism, and efficiency.

\smartparagraph{Proposed approach.}
We introduce \sys{}, an AI-agent framework that separates \emph{what} a workflow should do from \emph{how} it is executed. 
Users provide natural-language state prompts—high-level \emph{trigger--action} rules (e.g., ``if on login page, enter credentials,'' ``if viewing profiles, select \texttt{snlclient}'', ``if cookies popup, select \texttt{Accept}'')—which specify an abstract, non-linear workflow. 
A \emph{State Synthesis} component compiles these abstract prompts into \emph{concrete states} with application-bound detectors (DOM/text/URL) and reusable executable code. 
Concrete states are cached in a repository and deterministically replayed by a \emph{State Executor}; when UIs change, \sys{} regenerates only the affected states from the same abstract prompts. 
This compile–then–replay design blends the flexibility of language-based synthesis with the efficiency and stability of compiled execution, directly addressing the six requirements above.

\smartparagraph{Contributions and evidence.}
We implement \sys{} and evaluate it across 50+ workflows spanning video-on-demand streaming, live video streaming, video conferencing, social media, and web scraping (similar to BQT). 
Section~\ref{sec:design} details the abstractions and execution model; Section~3 demonstrates that (i)~abstract user prompts in natural language expand into hundreds of lines of executable code across diverse applications (extensibility), (ii)~caching and replay reduce token cost and make millions of repeat runs economically feasible (efficiency and repeatability), and (iii)~UI drift is handled by regenerating only the impacted state (robustness). 
Together, these results position \sys{} as a proof of concept for scalable and realistic data generation in networking, complementing controllable platforms such as netReplica~\cite{netreplica}---capable of emulating a diverse range of realistic networking conditions. We make \sys{} and its generated workflows available at \url{https://github.com/SNL-UCSB/netgent}.

\section{System Design}
\label{sec:design}


\label{sec:goals}

\subsection{Architectural Abstractions}
\label{sec:abstractions}
\sys{} separates \emph{what} a workflow should do from \emph{how} it is executed through three abstractions.

\smartparagraph{Abstract NFA.}
Users define an abstract nondeterministic finite automaton (NFA)~\cite{nfa} using natural-language state prompts. 
Each state prompt specifies \emph{triggers} (conditions that identify the state), \emph{actions} (intended task), and an optional \emph{end condition}. This representation captures non-linear flows (complexity) while keeping intent decoupled from UI specifics (robustness). 
For example, in a Disney+/ESPN workflow, states may include \texttt{login}, \texttt{select\_profile}, \texttt{navigate\_to\_espn}, \texttt{select\_video}, and \texttt{playback}.

\smartparagraph{Concrete NFA.}
During execution, \sys{} compiles each abstract state into a \emph{concrete state} defined by \(\hat{s}=(\textit{detectors},\textit{code})\): a set of CSS element, text, or URL detectors bound to the current application version, together with reusable executable code. This compiled form enables deterministic replay (repeatability) and cross-run reuse (efficiency). For example, the abstract trigger ``if on login page'' becomes a detector set (form labels, button text, stable DOM paths) and a short program that types credentials and clicks ``Log In.''

\smartparagraph{Cache and Replay.}
Concrete states are stored in a \emph{State Repository}; a \emph{State Executor} replays their code deterministically. If a detector later fails due to UI drift, only that state is regenerated from the abstract rule (robustness). Common states (e.g., \texttt{login}, \texttt{select\_profile}) are reusable across workflows and apps (efficiency, diversity).

\subsection{Workflow Execution Model}
\label{sec:workflow}
\begin{figure}[H]
    \centering
    \includegraphics[width=.9\linewidth]{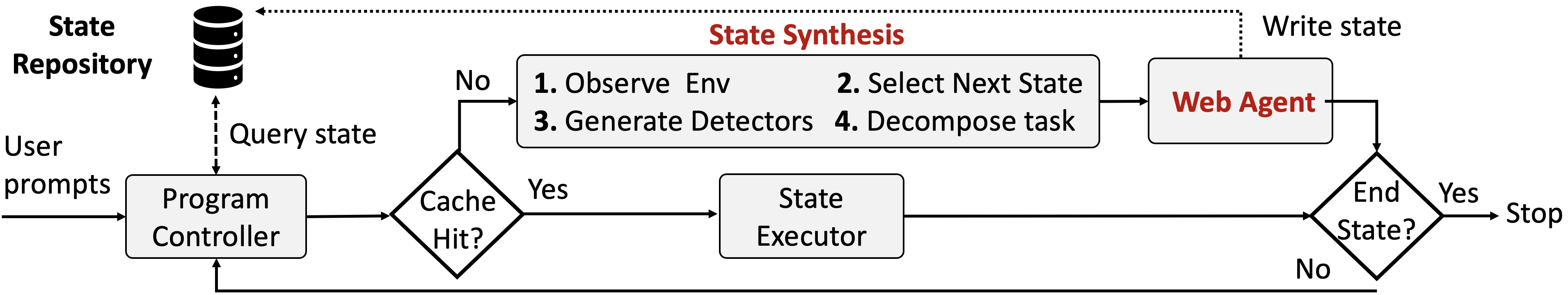}
    \caption{\sys{} runtime loop progresses from the initial to the end state}
    \label{fig:workflow}
\end{figure}

Figure~\ref{fig:workflow} illustrates the runtime loop which generates executable code from user prompts.

\smartparagraph{Controller queries the cache first.}
Given the current page (DOM) and the last transition, the \emph{Program Controller} queries the \emph{State Repository}. 
If a cache hit occurs, the Controller invokes the \emph{State Executor} to replay the stored code in the browser and the workflow advances.
This cache-first policy is the core of compile–then–replay and eliminates repeated reasoning (repeatability, efficiency).

\smartparagraph{Cache miss triggers one-shot synthesis.}
On a cache miss, the Controller invokes \emph{State Synthesis} (LLM), which performs four steps using the current DOM, screenshot, and user's rules:
(1) \emph{Observe} the environment to form a structured view;
(2) \emph{Select} the appropriate next abstract state (trigger–action pair);
(3) \emph{Generate} concrete detectors that reliably recognize that state;
(4) \emph{Decompose} the action into a simplified plan with decomposed tasks. The \emph{Web Agent} executes these decomposed tasks and generates the executable code. The \emph{Concrete State} is then written back to the repository. Only the missing node is synthesized; the abstract NFA and prior states remain intact (robustness, efficiency).

\smartparagraph{Realistic execution and termination.}
To enhance realism and evade bot detection, our web agent integrates browser stealth, human-like interaction, and network stealth (details in Appendix \S\ref{sec:realism}).
An end state is declared when an application-level condition holds (e.g., for ESPN, a \texttt{<video>} element is playing and time is advancing). Otherwise, the Controller loops to the next state.

\smartparagraph{Concrete example.}
Starting at the Disney+ homepage, the Controller hits cached \texttt{login} and \texttt{select\_profile} states on subsequent runs; on the first run these are synthesized once. Navigating to the ESPN hub and clicking the first video may trigger ads or a PIN prompt; the NFA branches handle these cases by synthesizing (once) a \texttt{type\_pin} or \texttt{skip\_ad} state and writing them to the repository. Playback detection serves as the end state, after which \sys{} records the successful trace and terminates.

\section{Evaluation}
We evaluate \sys{} against the requirements introduced in \S\ref{sec:goals}, mapping experiments to the abstractions in \S\ref{sec:abstractions} using Gemini 2.5 Flash \cite{Gemini25Our} with a temperature of 0.2. Details of the APIs and frameworks employed are provided in \S\ref{sec:lang}. All evaluations and executions were conducted on a  MacBook Pro (Apple M3 Pro chip, 11-core CPU, 14-core GPU, and 18~GB RAM).


\smartparagraph{Diversity across applications.}
A central goal of \sys{} is to keep user effort low while scaling to diverse applications. 
We hypothesize that prompt length serves as a proxy for user effort, while the lines of generated code reflect the automation complexity that would otherwise need to be implemented manually. 
We therefore measured the size of user prompts and the length of generated Python code across 50+ workflows spanning five domains: video-on-demand streaming, live video streaming, video conferencing, social media, and web scraping. Table~\ref{tab:token_costs} in the appendix provides the list of these applications, along with their evaluation based on code generation time, number of tokens used, and dollar cost.
Each workflow requires only a 100–200 word prompt to generate code that spans hundreds of lines.
This large expansion factor demonstrates that small, uniform specifications suffice to produce substantial executable workflows. 
Moreover, the same prompt structure generalizes across platforms within a category (e.g., Hulu and Disney+), showing that \sys{} is easily extensible to new applications with minimal effort.

\smartparagraph{Efficiency and repeatability.}
We next examine efficiency and repeatability within a specific workflow. The system leverages the compile–then–replay method to achieve repeatability by reusing stored concrete states, while caching reduces token usage by avoiding redundant LLM calls.  

We focus on the ESPN workflow, where user interactions include \texttt{login}, \texttt{select\_profile}, \texttt{playback} and other related actions. Running this workflow without any stored concrete NFA consumes 278k tokens per run, translating to \$0.098 at \$0.35 per million tokens\footnote{The total token cost is based on both input and output tokens, each typically charged at a different rate. For simplicity, we represent it here as the total number of tokens.}. Executing this workflow one million times without reusing a concrete NFA would cost roughly \$98,000 in LLM usage alone. By contrast, the compile–then–replay approach reuses stored states, eliminating per-run LLM costs and ensuring deterministic execution. This avoids redundant LLM and API calls, allowing a single generated workflow to be reused across multiple runs.

However, due to UI changes and other factors, periodic updates are necessary to handle new or modified states. Assuming a system with 10 states, generating each new state requires $\approx 42.6$k tokens ($\approx \$0.015$) on average. Generating workflows for all 10 states from scratch incurs a one-time cost of \\$\approx\$0.15$. If the workflow drifts weekly over a year (52 weeks), updating all states would cost $\approx \$7.8$.
With caching, only changed states need updating. Assuming one state update per week, the annual LLM cost drops to $\approx \$0.78$. Caching thus minimizes redundant LLM calls and enables deterministic replay of previously synthesized states. This demonstrates that compile–then–replay with state caching ensures deterministic behavior while making large-scale execution economically viable.

\smartparagraph{Robustness under UI drift.}
Finally, we evaluate robustness to interface changes. The hypothesis is that \sys{} can localize regeneration to only the affected state, avoiding costly re-synthesis of the full workflow. We perturb the ESPN workflow, requiring a PIN for profile access. In this case, only the affected state (\texttt{type\_pin}) was regenerated, while other states such as \texttt{login}, \texttt{select\_profile} and \texttt{navigate\_to\_espn}  were replayed from cache without modification. Using the caching method, regenerating the affected state required only $\approx 20$k tokens and the entire process took 216 seconds, compared to $\approx 375$k tokens and 406 seconds if done from scratch. This bounded overhead confirms that state-level regeneration is sufficient. Even when multiple states change, the unaffected portions of the workflow remain intact. Such robustness ensures that workflows remain usable despite frequent UI drift in production applications. See Figure~\ref{fig:espn} in the appendix for the full ESPN workflow. 

\smartparagraph{Summary.}
Across all experiments, \sys{} satisfies the requirements of \S\ref{sec:goals}: prompts abstract away application-specific details to ensure diversity and extensibility; caching enables efficiency and repeatability; and state-local regeneration ensures robustness to UI drift. 
These results collectively demonstrate that \sys{} provides a scalable foundation for generating realistic networking datasets across heterogeneous applications.

\section{Limitations and Future Work} 
While \sys{} demonstrates that abstract NFAs combined with compile--then--replay enable scalable and repeatable workflows, several limitations remain. First, manual workflow verification and failure handling currently require user intervention; automating step-level validation and state-level recovery would enable self-healing and fully autonomous workflows. Second, \sys{} is limited to web applications; extending the NFA abstraction to desktop environments would broaden applicability. 
Together, these extensions will make \sys{} more autonomous, robust, and broadly deployable.

\section*{Acknowledgement}
This work was supported in part by the National Science Foundation (CAREER Award No. 2443777 and CNS Award No. 2323229) and a research gift from Cisco and Google. Eugene Vuong was supported by the Cal-Bridge Program. This research used resources of the National Energy Research Scientific Computing Center, a DOE Office of Science User Facility supported by the Office of Science of the U.S. Department of Energy under Contract No. DE-AC02-05CH11231 using NERSC award NERSC DDR-ERCAP0029768.

\bibliographystyle{plain}
\bibliography{ref}

\medskip

{
\small


\appendix
\section{Appendix / Supplemental Material}

\begin{figure}[H]
    \centering
    \includegraphics[width=\linewidth]{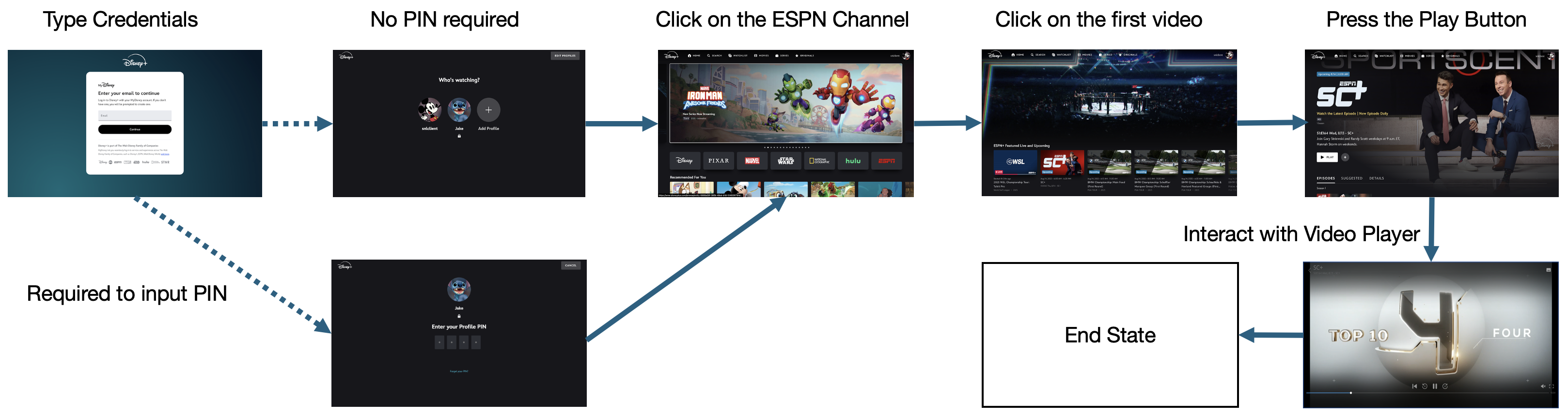}
    \caption{ESPN workflow: log into Disney+, select the account, enter the PIN if required, navigate to ESPN, play the first video, and advance the playback slider to the five-minute mark.}
    \label{fig:espn}
\end{figure}


\begin{table}[ht]
\centering
\begin{threeparttable}
\caption{Evaluation of \sys{} across different workflows, performed entirely without using cached states.
}
\label{tab:token_costs}
\begin{tabular}{lllll}
\hline
Application Type & Platform & Tokens ($\times 10^3$) & Price (\$) & Time (s) \\
\hline
Video Conferencing & Zoom~\cite{zoom}       & 146.8 & 0.051 & 138 \\
Video Conferencing & Microsoft Teams~\cite{microsoft_teams}  & 242.9 & 0.085 & 238 \\
Video Conferencing & Google Meet~\cite{google_meet}     & 169.7 & 0.060 & 160 \\
Video Conferencing & Zoho Meeting~\cite{zoho_meeting}     & 137.6 & 0.049 & 190 \\
Video Conferencing & Jitsi~\cite{jitsi}            & 98.5  & 0.035 & 120 \\
Video Conferencing & Whereby~\cite{whereby}          & 77.1  & 0.026 & 78  \\
Video Conferencing & Ring Central~\cite{ringcentral}     & 131.9 & 0.047 & 139 \\
Video Conferencing & Talky~\cite{talky}            & 110.9 & 0.041 & 129 \\
Video Conferencing & Webex~\cite{webex}            & 107.0 & 0.037 & 114 \\
\hline
Video on Demand + Live stream & YouTube~\cite{youtube}      & 131.7 & 0.047 & 105 \\
Video on Demand + Live stream & ESPN~\cite{espn}         & 278.7 & 0.098 & 339 \\
Video on Demand + Live stream & ESPN (PIN Required)         & 375.4 & 0.105 & 406 \\
Video on Demand + Live stream & ESPN (PIN Required)$^{\dagger}$        & 20.1 & 0.006 & 216 \\
\hline
Video on Demand & Disney Plus~\cite{disneyplus}       & 249.2 & 0.088 & 335 \\
Video on Demand & Hulu~\cite{hulu}              & 245.0 & 0.089 & 386 \\
Video on Demand & Roku~\cite{roku}              & 162.2 & 0.059 & 187 \\
Video on Demand & Tubi~\cite{tubi}              & 140.6 & 0.049 & 158 \\
\hline
Live stream & Twitch~\cite{twitch}        & 134.6 & 0.044 & 90  \\
Live stream & Puffer~\cite{puffer_stanford}        & 109.3 & 0.038 & 126 \\
\hline
Social Media & X (Twitter)~\cite{x}  & 201.7 & 0.068 & 158 \\
Social Media & Instagram~\cite{instagram}    & 199.1 & 0.069 & 180 \\
Social Media & LinkedIn~\cite{linkedin}     & 143.8 & 0.049 & 138 \\
Social Media & Reddit~\cite{reddit}       & 739.5 & 0.225 & 82  \\
Social Media & Bluesky~\cite{bluesky}      & 98.1  & 0.033 & 79  \\
\hline
Web Scraping & BQT~\cite{bqt} (30 ISPs)      & 145* & 0.050* & 120* \\
\hline
\end{tabular}
\begin{tablenotes}
\small
\item[$\dagger$] This execution uses the cached ESPN workflow (no PIN) and shows \sys{}’s efficiency under UI drift.
\item[$*$] Approximate average numbers across 30 ISPs.
\end{tablenotes}
\end{threeparttable}
\end{table}

\subsection{Language and API Integration}
\label{sec:lang}

Our system’s architecture is built by composing several key technologies. The foundational layer for interacting with LLMs is standardized using core LangChain \cite{LangChain} abstractions, such as \texttt{SystemMessage} and \texttt{HumanMessage} for model inputs and tool invocation. We specifically access Google's Gemini 2.5 chat models \cite{Gemini25Our} through the VertexAI platform, which is integrated as a backend via the \texttt{langchain-google-vertexai} wrapper. While LangChain provides the core communication components, we use LangGraph as the control-flow layer to orchestrate the Browser Agent's complex behavior. LangGraph allows us to define the agent's logic, which is essential for managing its multi-step processes of environment observation, planning, and code generation and execution.

To enhance the Web Agent’s capabilities and efficiency, we incorporated several key open-source contributions. Notably, we integrated the DOM Marker code from Browser Use \cite{browser_use2024} to implement Set-of-Mark (SoM) prompting \cite{yangSetofMarkPromptingUnleashes2023}, enabling precise interaction with web elements via visual markers. We also adopted the Planner, Replanner, and Executor prompts from Plan-and-Act \cite{erdoganPlanandActImprovingPlanning2025}, allowing the agent to decompose complex tasks into structured planning phases followed by systematic execution. These integrations provide a robust foundation that leverages proven open-source techniques while supporting the unique requirements of \sys{}, facilitating reliable and scalable automation of complex web workflows.

\subsection{Realism: Mimicking Human-Like Interactions}
\label{sec:realism}
To support realism, we adopted multiple techniques in the web agent so tasks simulate human-like web interactions and avoid bot detection mechanisms. To mitigate automated-behavior detection, our system integrates three anti-bot methods: \emph{(i) browser stealth}, \emph{(ii) movement realism}, and \emph{(iii) network stealth}. For \emph{browser stealth}, we use SeleniumBase \cite{seleniumbase} with undetected-chromedriver \cite{undetected-chromedriver} to hide common automation fingerprints. Besides that, repeated logins during automated workflows can trigger service defenses, such as account freezes or forced two-step verification. To overcome this issue, we support persistent user profiles via the \texttt{user-data-dir} option. This command makes the browser retain state across sessions, including cookies (login information) and cached site assets. This prevents repeated login flows and creates a continuous session experience similar to a regular user. Combined with fingerprinting avoidance, this supports a stable, persistent, and human-like browser profile.

To incorporate \emph{movement realism}, we use human-like interaction primitives. Instead of modifying the page DOM directly, the system computes absolute screen coordinates and issues mouse and keyboard events through PyAutoGUI \cite{pyautogui}. Mouse movements follow smooth Bezier-curve trajectories~\cite{bezier} rather than instant jumps. Keystrokes are generated with small, variable delays. Scrolling and hovering are restricted to the visible viewport, avoiding sudden jumps and ensuring the agent only interacts with elements a human user can actually see.

Finally, we provide \emph{network stealth} by using a pool of IP addresses from Bright Data~\cite{BrightDataAll} to distribute requests across different network origins. This is essential for many web scraping applications~\cite{bqt,bqtplus}, as repeated interactions from a single IP increase the likelihood of rate limits or blocking. By varying the network origin, the system reduces such detection signals while maintaining the appearance of normal user traffic. While effective in these settings, this approach is less suitable for services like video streaming that rely on IP-based account tracking to prevent oversharing. Together, these methods allow the system to closely mirror human interactions with applications and the browser, which is critical for generating robust workflows that can be executed repeatedly and reliably without triggering blocking mechanisms.

\subsection{Broader Use Cases of \sys{} for Networking Data Generation}
The use cases and importance of \sys{} extend beyond generating traffic for the control application targeted for data collection. A high-quality, generalizable dataset must be collected in diverse network environments, where one of the most challenging aspects to emulate is cross traffic: Replace it with: traffic from other applications that share the same path or bottleneck link and influence the control application, such as inducing congestion events or triggering ABR algorithms
~\cite{congestion,abr}. The main challenge is generating cross traffic that is realistic, diverse, and reactive~\cite{netrep_new}. A practical approach is to construct a large pool of diverse application workflows and select combinations of them to run alongside the control application. This allows us to emulate realistic environments, such as a home network where one user is streaming video while another is browsing, and evaluating how the control application (e.g., Zoom) behaves under these shared and congested network conditions.




\end{document}